\documentclass[10pt, a4paper]{article}
\usepackage{lrec2022} % this is the new LREC2022 Style
\usepackage{multibib}
\newcites{languageresource}{Language Resources}
\usepackage{graphicx}
\usepackage{tabularx}
\usepackage{soul}

\newcommand{\T}{\ensuremath{\mathcal{T}}\xspace}

\usepackage{times}
\usepackage{graphicx} \usepackage{grffile}

\usepackage{booktabs}
\usepackage{longtable}
\usepackage{amssymb}
\usepackage{makecell}
\usepackage{amsmath}
\usepackage{xspace}
\usepackage{scrextend}

\usepackage{titlesec}
\titleformat{\section}{\normalfont\large\bfseries\center}{\thesection.}{1em}{}
\titleformat{\subsection}{\normalfont\SmallTitleFont\bfseries\raggedright}{\thesubsection.}{1em}{}
\titleformat{\subsubsection}{\normalfont\normalsize\bfseries\raggedright}{\thesubsubsection.}{1em}{}
\renewcommand\thesection{\arabic{section}}
\renewcommand\thesubsection{\thesection.\arabic{subsection}}
\renewcommand\thesubsubsection{\thesubsection.\arabic{subsubsection}}
\usepackage{epstopdf}
\usepackage[utf8]{inputenc}

\usepackage{hyperref}
\usepackage{xstring}

\usepackage{color}

\title{Analysis and Prediction of NLP models via Task Embeddings }

\name{Damien Sileo, Marie-Francine Moens} 

\address{KU Leuven\\
         Belgium \\
         damien.sileo@kuleuven.be, sien.moens@cs.kuleuven.be}

\abstract{
 Task embeddings are low-dimensional representations that are trained to capture task properties.  In this paper, we propose MetaEval, a collection of $101$ NLP tasks. We fit a single transformer to all MetaEval tasks jointly while conditioning it on learned embeddings. The resulting task embeddings enable a novel analysis of the space of tasks. We then show that task aspects can be mapped to task embeddings for new tasks without using any annotated examples. 
 Predicted embeddings can modulate the encoder for zero-shot inference and outperform a zero-shot baseline on GLUE tasks. The provided multitask setup can function as a benchmark for future transfer learning research. 
 \\ \newline  }

\usepackage{microtype}
\newcommand{\blue}[1]{\textcolor{black}{#1}}

\begin{document}

\maketitleabstract

\section{Introduction}

Transfer between tasks enabled considerable progress in NLP. Pretrained transformer-based encoders, such as BERT \cite{devlin-etal-2019-bert} and RoBERTa \cite{liu2020RoBERTa}, achieved state-of-the-art results on text classification tasks. These models acquire rich text representations through masked language modeling (MLM) pretraining \cite{tenney-etal-2019-bert,Warstadt2019InvestigatingBK,warstadt-etal-2020-learning}. However, these representations need additional task supervision to be useful for downstream tasks \cite{reimers-2019-sentence-bert}. 
The default technique, {\it full fine-tuning}, optimizes all encoder weights alongside the training of the task-specific classifier.
The resulting encoder weights %$\theta_{\mathcal{T}}$ 
can be seen as a very high-dimensional\footnote{E.g., $\approx 110M$ dimensions for $\text{BERT}_{\text{BASE}}$ full fine-tuning.} continuous representation of a model that is dedicated to a task $\T_i$ \cite{aghajanyan2020intrinsic}. %As such, we can consider them as a representation of the target task $\T_i$.
These weights provide a way to predict relatedness between tasks
\cite{achille2019task2vec,vu-etal-2020-exploring}. 
\blue{However, they are very high dimensional, and they cannot be  modulated to adapt the network to unseen tasks.}
Hypernetworks \cite{Ha17,Oswald2020Continual,Hansen2020Fast}, i.e., neural networks whose weights are modulated by an outer network, can solve this problem. These techniques have  been adapted to NLP by \newcite{CAMTL2021} and \newcite{mahabadi2021parameterefficient} who rely on adapters  \cite{HoulsbyGJMLGAG19}. Adapters are parameter-efficient layers that can be inserted between specific layers and trained to modulate a frozen transformer. An adapter $A_i$ is composed of distinct adapter layers. 
\newcite{CAMTL2021} and \newcite{mahabadi2021parameterefficient} showed that in a multitask setting with a collection of tasks $\Theta$, a set of adapters $\{ A_i, \T_i \in \Theta \}$ can be decomposed into two components: a set of task embeddings $\{z_i, \T_i \in \Theta \}$ and a single shared conditional adapter $A(z_i)$. The task embeddings and conditional adapter are trained jointly, which allows each task to modulate the shared model in its own way. This approach leads to a performance improvement over individual adapters or full fine-tuning while allowing very low-dimensional  ($\text{dim(z)} <  100$)  task representations.% =\theta_{\mathcal{T}} $. 

In this work, we leverage conditional adapters to derive task embeddings for $101$ tasks based on a joint multitask training objective. This enables new  analyses of the relationships among the tasks. We show that we can predict the task embeddings from selected task aspects, which leads to a more selective and interpretable control of NLP models.

We answer the following research questions:
RQ1: How consistent is the structure of task embeddings? What is the importance of weight initialization randomness and sampling order on a task embedding position within a joint training run? How similar are task relationships across runs? RQ2: A consistent structure allows meaningful probing of the content of task embeddings. How well can we predict aspects of a task, such as the domain, the task type, or the dataset size, based on the task embedding? RQ3: Task embeddings can be predicted from task aspects, and a task embedding modulates a model. Can we predict an accurate model for zero-shot transfer based solely on the aspects of a task?

Since we study task representations, many tasks and, ideally, many instances for each task type are required for our analysis. Consequently, we have assembled $101$ tasks in a benchmark that can be used for future probing and transfer learning. 
Our contributions are the following:
(i) We assess low-dimensional task embeddings in novel ways, enabling their in-depth analysis;
(ii) We show that these embeddings contribute to transferring models to target downstream NLP tasks even in situations where no annotated examples are available for training the downstream NLP task;
(iii) We introduce MetaEval, a benchmark framework containing $101$ NLP classification tasks\footnote{\url{https://github.com/sileod/metaeval}}.

\begin{figure}
  \centering
  \vspace{-0.2cm}
%\includegraphics[trim={0 1.2cm 2.0cm 0},clip,
%width=0.5\textwidth]{figures/slides12c.pdf}
\includegraphics[width=0.5\textwidth]{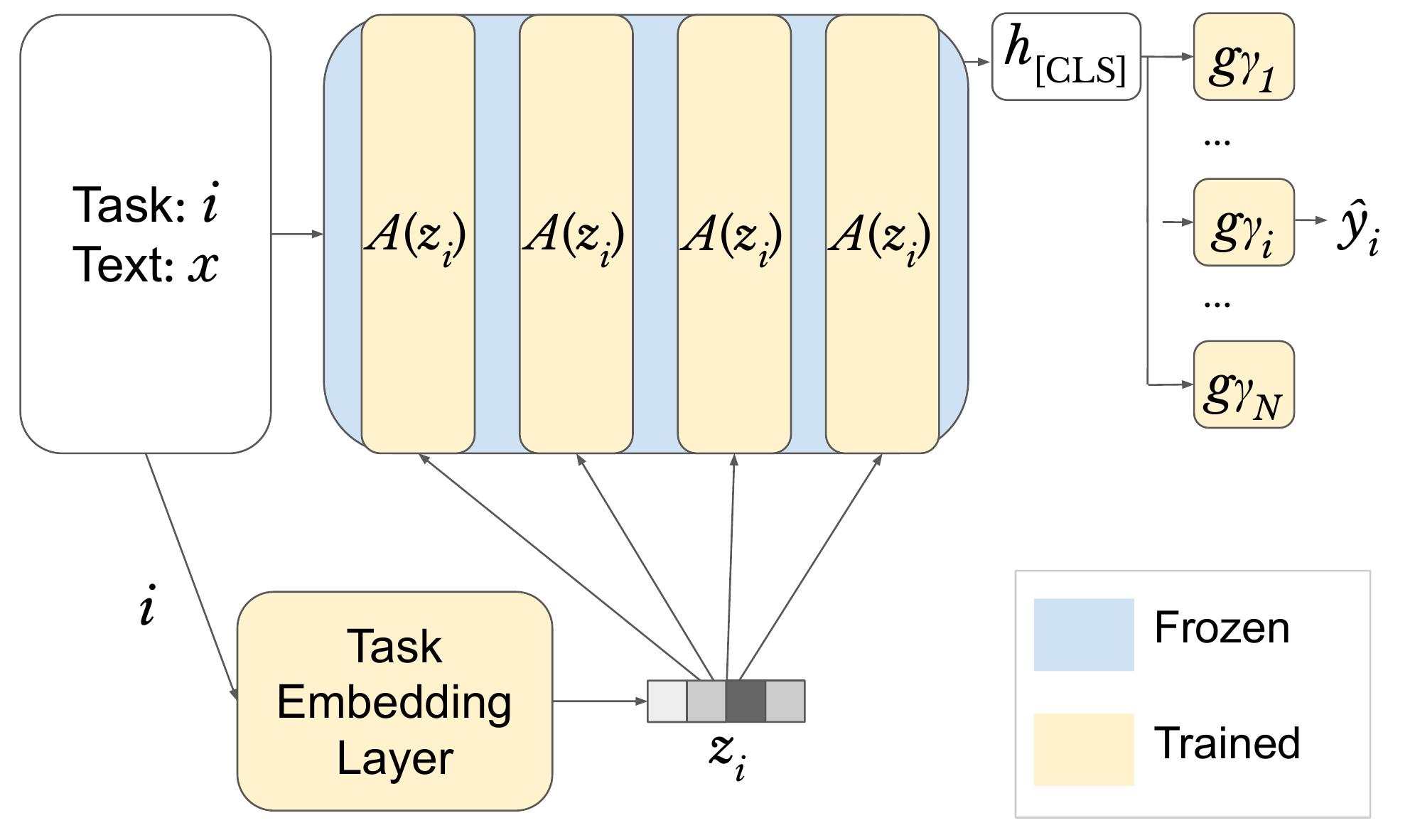}
  \caption{An overview of a transformer with a conditional adapter in a classification setup with $N$ tasks. Batches for each task are used sequentially in random order. Each text example $x$ is represented by $h_{\text{[CLS]}}$, which is the input of $g_{\gamma_i}$ and the classifier for the task $\T_i$. }
\label{fig:overview} 
\end{figure}

\section{Related Work}

\paragraph{Task relatedness and task embeddings} A common way to measure task relatedness is to train a model on a source task, or a combination of source tasks in the case of multitask learning  \cite{Caruana1997}, and then measure the effect on the target task's accuracy.
The search for the most useful source tasks for each target task has been the object of numerous studies. \newcite{mou-etal-2016-transferable} study the effect of transfer learning when the target task has a different domain from the source task and focus on different fine-tuning strategies, for instance, freezing or unfreezing specific layers.
\newcite{conneau-etal-2017-supervised} train a sentence encoder with a selection of source tasks and show that natural language inference (NLI) provides the most transferable representations.
\newcite{phang2018stilts} also address the fine-tuning of pretrained BERT with a two-stage approach: an auxiliary pretraining stage on a source task before the final fine-tuning on the target task.
%\newcite{Wu2020Understanding} investigate the phenomenon of negative transfer, i.e., the situation where source tasks harm target tasks in a multitask setting, and propose techniques to alleviate this phenomenon.
\newcite{d2020underspecification} show that when fine-tuning a model for a task, various random seeds can lead to similar accuracy but different behavior on subtasks. We perform a comparable analysis in a multitask setup and show that task embeddings are a valuable way to visualize this phenomenon. By contrast, we do not study the transferability of task on each other, but we evaluate the properties of tasks in the latent space.
%prompt fine-tuning (R2)
Task embeddings were formalized and linked to task relatedness in computer vision tasks by \newcite{achille2019task2vec}, who interpret pooled Fisher information in convolutional neural networks as task embedding. They treat each label as a task and compare task embeddings with labels. \newcite{vu-etal-2020-exploring} adapt this task embeddings technique to NLP models but they limit their analysis that to the prediction of task relatedness. Here, we also evaluate Fisher embeddings in the NLP context but also compare them to conditional adapter embeddings and probe task properties.

\paragraph{Probing neural text representations} Our work is also related to the probing of representations, which usually targets words \cite{nayak-etal-2016-evaluating} or sentences. \newcite{conneau-etal-2018-cram} probe sentence representations for various syntactical and surface aspects. Another type of probing,  proposed for word embeddings, is the study of stability \cite{pierrejean-tanguy-2019-investigating,antoniak-mimno-2018-evaluating,wendlandt-etal-2018-factors}. Stability measures the similarity of word neighborhoods across different training runs with varying random seeds. 

\paragraph{Transfer techniques} Several alternatives were proposed to overcome the shortcomings of full fine-tuning. \newcite{HoulsbyGJMLGAG19} proposed adapters as a compact transformation to modulate a model without fine-tuning the whole network.  \newcite{pmlr-v97-stickland19a} leverage adapters in a multi-task setting with a fine-tuned transformer and task-specific adapters. \newcite{CAMTL2021} and  \newcite{mahabadi2021parameterefficient} modulate a single adapter with task embedding to enable efficient multi-task learning and compact task representation, but do not perform inference on new tasks.  \newcite{Oswald2020Continual} propose a task inference model based on input data for continual learning problems on vision tasks, and \newcite{Hansen2020Fast} also applies this idea reinforcement learning for visual tasks. \newcite{cao2020modelling} address language generation on a variety of domains, which can be treated as tasks. They also rely on input data to predict task embeddings. Here, we adapt the idea of task inference from input data to NLP classification tasks, but we also show that known task attributes such as task type can be used instead of the input data. This is analogous to \newcite{ustun-etal-2020-udapter} who use typological language features for adaptation of dependency parsing to new languages. Finally, prompts can be also used for transfer without fine-tuning  \cite{radford2019language} or by tuning token embeddings to learn a prompt \cite{li2021prefixtuning,qin-eisner-2021-learning}, but they are used for text generation or knowledge probing which are outside the scope of this work.

\section{Models and Setups}
\label{sec:models}
We now introduce the classification models and fine-tuning techniques used in our experiments. To perform a classification task $\T_i$, we represent a text $x$ (e.g., a sentence or a sentence pair) with an encoded \textsc{[CLS]}  $d$-dimensional token  $h_\text{[CLS]}=f_{\theta}(x)$. Here, $f_{\theta}$ is a transformer text encoder.  $h_\text{[CLS]}$ is used as the input features for a classifier $g$.  For each task, we use a different classification head $g_{\gamma_i}$, where $\gamma_i$ represents softmax weights. To train a model for a task, we minimize the cross-entropy $\text{H}(y_i, g_{\gamma_i}(f_{\theta}(x)))$ where $y$ denotes a label. Different strategies can be used to fine-tune a pretrained text encoder $f_{\theta_{\text{MLM}}}$ for a set of tasks:

\paragraph{Full Fine-Tuning} is the optimization of all parameters of the transformer architecture alongside classifier weights, ($\theta_i, \gamma_i)$, independently for each task.

\paragraph{Adapters} are lightweight modules with new parameters $\alpha$ that are inserted between each attention and feed-forward transformer layer \cite{HoulsbyGJMLGAG19}. When using adapters ($ A_{\alpha_i}$), we freeze the transformer weights and represent each input text as $h_\text{[CLS]}=f_{\theta_{\text{MLM}}, A_{\alpha_i}}(x)$.  During adapter fine-tuning, we optimize only the adapter weights and classifier weights ($\alpha_i, \gamma_i $) for each task.

\paragraph{Conditional Adapters} We replace task-specific adapters with conditional adapters $A_\alpha(z_i)$ that are common to all tasks but conditioned on task embeddings $z_i$. To do so, we train all the tasks jointly and optimize a conditional adapter that learns to map each task embedding to a specific adaptation of the transformer weights while simultaneously optimizing the task embeddings.  Figure \ref{fig:overview} shows an overview of our conditional adapter setup. The objective is the following:
\[
%\mathcal{L}_{\text{multitask}}=
\text{min}_{(\alpha,z,\gamma)}
\sum_{\T_i \in \Theta} \text{H}(y_i, \hat{y}_{i})
\]
\begin{figure}
  \centering
  \vspace{-0.2cm}
\includegraphics[%trim={0cm 2cm 15cm 0},clip,
width =0.42\textwidth]{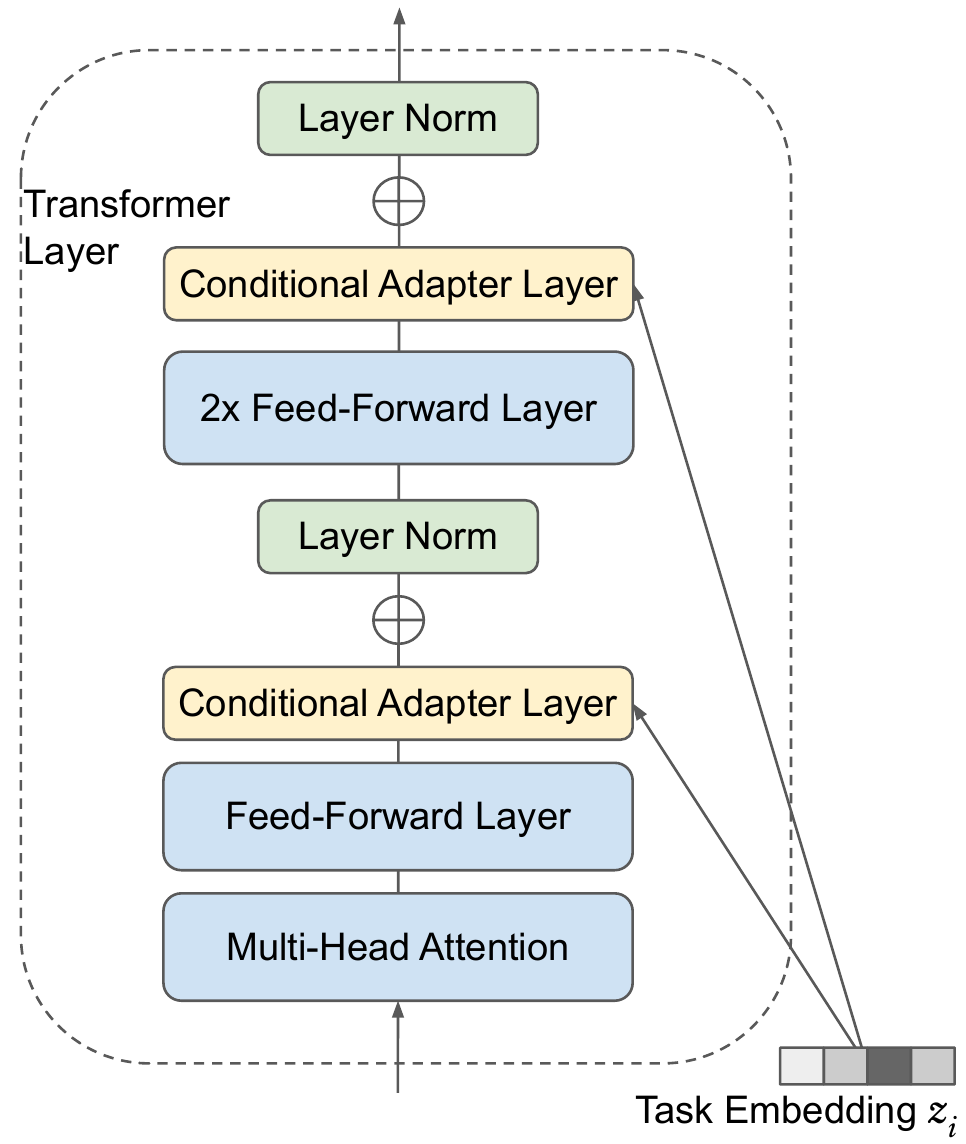}
  \caption{A transformer layer with conditional adapter layers. \label{fig:hyperadapter}}
\end{figure}

\subsection{Parametrization of Adapters and Conditional Adapters}
%Here, we provide more details on the parametrization of Adapters and Conditional Adapters.

Figure \ref{fig:hyperadapter} illustrates two conditional adapter layers in a transformer layer.
An adapter layer is a one hidden layer perceptron with a bottleneck of dimension $a$. Each adapter layer applies the following transformation:
\begin{equation}
h \rightarrow h+\text{LayerNorm}_{\gamma ,\beta}( U(\text{GeLU}(D(h)))
\end{equation}

where $D$ and $U$ are linear down-projection and up-projection matrices in $\mathbb{R}^{d\times a}$ and $\mathbb{R}^{a\times d}$ respectively.
Adapter layers are inserted between fixed-weight transformer layers  to adjust the text representation for the target task. Layer normalization weights $\gamma, \beta$ \cite{ba2016layer} are also optimized and are considered as a part of the adapters.

%Conditional adapters \cite{CAMTL2021,mahabadi2021parameterefficient} are an extension of adapters designed for parameter efficiency in multitask setups. For all tasks, a conditional adapter is modulated by task-specific embeddings. 

In a conditional adapter, LayerNorm weights are modulated by $z$ in the following way: $\gamma, \beta = W_{\gamma}z, W_{\beta}z$ where $W_\gamma$ and $W_{\beta}$ are learnable randomly initialized projections in $\mathbb{R}^{\text{dim}(z) \times d}$.

\newcite{mahabadi2021parameterefficient} use a similar modulation of the $D,U$ matrices and generate their weight: $D, U = W_D z, W_U z $ with $W_D \in \mathbb{R}^{(d \times a) \times \text{dim(z)}}$, $W_U \in \mathbb{R}^{(a \times d) \times \text{dim(z)}}$. They also show that adapters can be shared across layers, but this did not lead to improvement in our experiments.

Instead, \newcite{CAMTL2021} use the following transformation\footnote{\newcite{CAMTL2021} also have proposed a conditional attention which did not yield improvement in our experiments}:
\begin{equation}
h \rightarrow  h + W_h z \odot h + W_b z
\end{equation} where $W_h$ and $W_b$ are projections in $\mathbb{R}^{\text{dim(z)}\times d}$, before each adapter layer.

%Tasks that are close in the task embedding space influence the feature extraction of the transformer in a similar way. Each layer has distinct conditional adapter weights, but a task embedding is shared across all layers.

\subsection{Baseline task embeddings}\label{sec:baselinetaskembeddings}
We also perform experiments with the task embeddings methods proposed by \newcite{vu-etal-2020-exploring}  instead of a learned task embedding. We  project them to $\text{dim}(z)$ with a randomly initialized trainable linear layer.
\paragraph{TextEmb} is the average text embedding across all examples of a task. We use the average of the output tokens \cite{vu-etal-2020-exploring,reimers-2019-sentence-bert} as text embeddings.
\paragraph{Fisher Embedding} captures the influence of the training objective on the activation of  $h_{\text{[CLS]}}$. See appendix \ref{sec:fisher} for additional details.

\section{Datasets}
One of our goals is to study and leverage the task embeddings by making use of known task aspects. This process involves a mapping between the task and the aspects, which requires a varied set of tasks. The most commonly used evaluation suite, GLUE, contains only 8  datasets, which is not sufficient for our purpose. Therefore, we construct the largest set of NLP classification tasks\footnote{We concentrate on English text classification tasks due to their widespread availability and standardized format.} to date by casting them into the HuggingFace Datasets library.

\paragraph{HuggingFace Datasets} \cite{2020HuggingFace-datasets} is a repository containing individual tasks and benchmarks including GLUE \cite{wang2018glue} and SuperGLUE \cite{wang2019superglue}. We manually select classification tasks that can be performed from single-sentence or sentence-pair inputs and obtain $39$ tasks.

%We also add the following benchmarks that were not in HuggingFace Datasets to build the MetaEval benchmark.

\begin{table*}

%baseline 0.2, none : 18.3, single: 26.1, gated : 15.03 %TODO
\centering
\small
\begin{tabular}{lrll}
\toprule
  Fine-Tuning Method &   \thead[l]{MetaEval\\ Test Accuracy} &  \thead[l]{Trained Encoder\\ Parameters} &  \thead[l]{Task Specific \\Trained Encoder Parameters} \\
\midrule
                Majority Class &                    42.9 &                       - &                             - \\
                Full-Fine-Tuning (1 model/task) &                    76.9 &                       124M &                             124M \\
             Adapter &                    67.8 &                        10M &                              10M \\
 Conditional Adapter \cite{mahabadi2021parameterefficient}&           75.6          &                        38M &                               512 \\
 Conditional Adapter \cite{CAMTL2021} &                     \textbf{79.7} &                        10M &                               32 \\
 \hspace{5mm}$z$=TextEmb task embedding \cite{vu-etal-2020-exploring} &           69.9 &                        10M &                               32 \\
\hspace{5mm}$z$=Fisher information task embedding \cite{vu-etal-2020-exploring} &           67.5 &                        10M &                               32 \\

\bottomrule
\end{tabular}
\caption{Parameter counts and MetaEval test accuracy percentages of fine-tuning techniques. The last two rows replace the latent task embedding $z$ with a linear projection of the task features proposed by \protect\newcite{vu-etal-2020-exploring}.}
\label{tab:metaevalacc}
\end{table*}

\paragraph{CrowdFlower} \cite{van2012designing} is a collection of datasets from the CrowdFlower platform for various tasks such as sentiment analysis, dialog act classification, stance classification, emotion classification, and audience prediction.

\paragraph{Ethics}\cite{hendrycks2021aligning} is a set of ethical acceptability tasks containing natural language situation descriptions associated with acceptability judgment under 5 ethical frameworks.
\paragraph{PragmEval}\cite{sileo2019discoursebased} is a benchmark for language understanding that focuses on pragmatics and discourse-centered tasks containing $23$ classification tasks.
\paragraph{Linguistic Probing} \cite{conneau-etal-2018-cram} is an evaluation designed to assess the ability of sentence embedding models to capture various linguistic properties of sentences with tasks focusing on sentence length, syntactic tree depth, present words, parts of speech, and sensibility to word substitutions. 
\paragraph{Recast} \cite{poliak-etal-2018-collecting-diverse} reuses existing datasets and casts them as NLI tasks. For instance, an example in a pun detection dataset  
\cite{yang-etal-2015-humor} \textit{Masks have no face value} is converted to a labeled sentence pair (\textit{Kim heard masks have no face value;
Kim heard a pun }y=\textsc{entailment})

\paragraph{TweetEval} \cite{barbieri2020tweeteval} consists of classification tasks focused on tweets. The tasks include sentiment analysis, stance analysis, emotion detection, and emoji detection.
\paragraph{Blimp-Classification} is a derivation of BLIMP \cite{warstadt-etal-2020-blimp-benchmark}, a dataset of sentence pairs containing naturally occurring sentences and alterations of these sentences according to given linguistic phenomena. We recast this task as a classification task, where the original sentence is acceptable and the modified sentence is unacceptable.

The table in Appendix \ref{taskslist} displays an overview of the tasks in MetaEval. When splits are not available, we use $20\%$
 of the data as the test set and use the rest for an $80/20$ training/validation split.
 We will make the datasets and splits publicly available.
 % (i.e., what column is the label column, what text columns should be used as first and second input) publicly available.

\section{Experiments}

Our first goal is to analyze the structure and regularity of task embeddings. We then propose and evaluate a method to control models using task aspects.

\subsection{Setup }\label{sec:setup}

%\paragraph{Conditional Transformer Hyperparameters}
Following \newcite{CAMTL2021}, we use a $\text{RoBERTa}_{\text{BASE}}$ \cite{liu2020RoBERTa} pretrained transformer\footnote{$\text{BERT}_{\text{BASE}}$ had a similar behavior in our experiments, but with a slightly lower accuracy. } , a sequence length of $128$, a batch size of $64$, and Adam with a learning rate of $2.10^{-5}$ as an optimizer during $3$ epochs for single-tasks model, 1 epoch while multitasking and an adapter size $a=256$ with \cite{CAMTL2021} and $a=32$ with \cite{mahabadi2021parameterefficient} as they suggest. We use the same hyperparameters for the baselines otherwise (tuning them did not lead to significant improvement). We set a limit of $30k$ training examples per task per epoch to obtain manageable computation time. 

\paragraph{Multitask setup} When multitasking, we sample one task from among all MetaEval tasks at each training step. The loss for each task is capped to $1.0$ to prevent unbalance between tasks. We also sample each task with a probability proportional to the square root of the dataset size \cite{pmlr-v97-stickland19a} to balance the mutual influence of the tasks.
We use task embeddings of  $\text{dim}(z)=32$, which was selected according to MetaEval average validation accuracy among $\{2,8,32,128,512\}$ and is also suggested by \newcite{mahabadi2021parameterefficient}.

\begin{figure*}

  \centering
  \vspace{-0.2cm}
\includegraphics[trim={4cm 3cm 3.5cm 3cm},clip,
width =0.9\textwidth,]{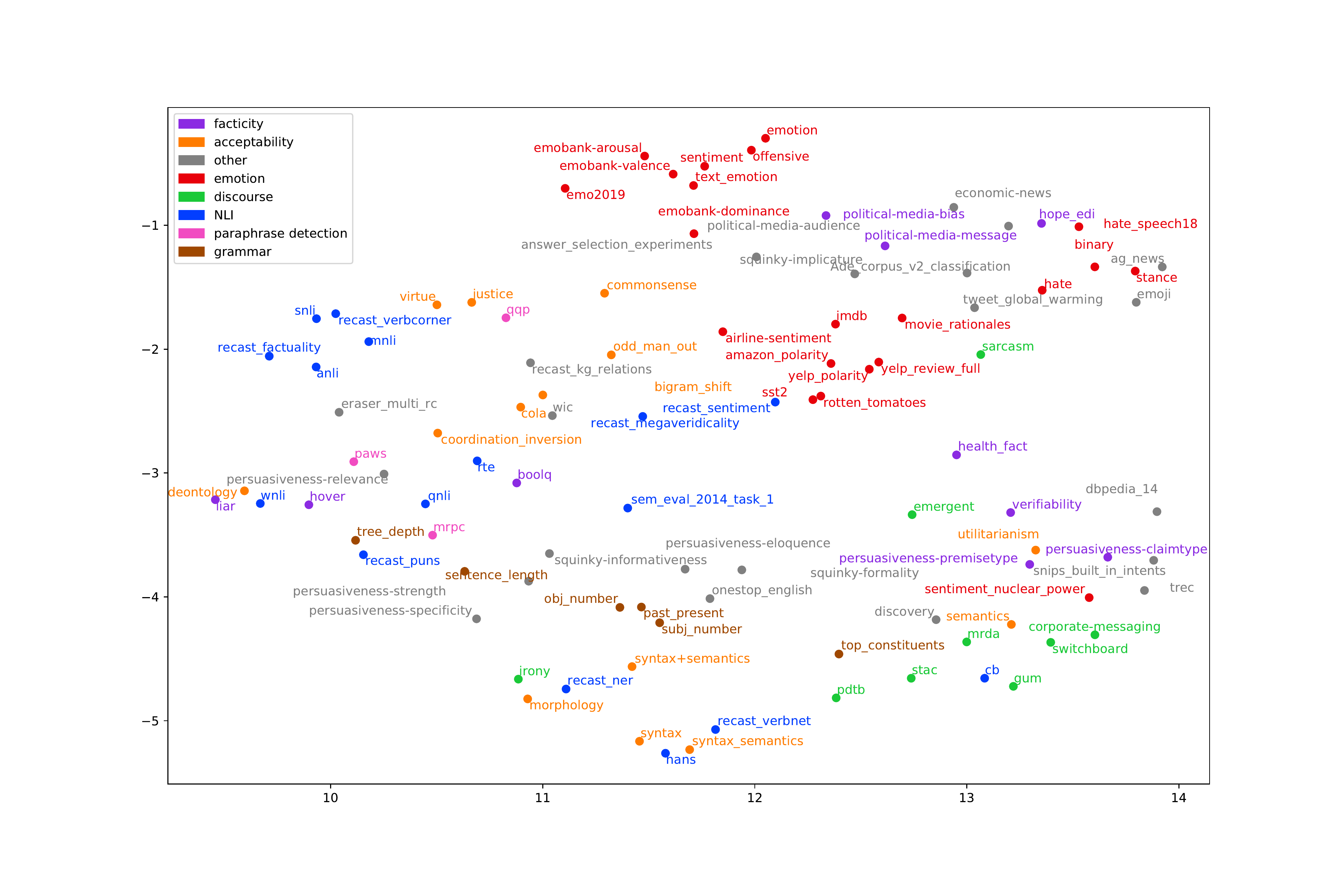}
  \caption{ UMAP projection of the task embeddings. }
\label{fig:taskviz} 
\end{figure*}
\subsection{Target Task Results \label{sec:targettask}}

We first evaluate the individual model performance  for the settings described in section \ref{sec:models}.

Table \ref{tab:metaevalacc} compares the unweighted average of the accuracies computed for MetaEval tasks and the number of trainable parameters associated with the fine-tuning strategies.
The conditional adapters achieve comparable accuracy to that of full fine-tuning despite having only 32 task-specific encoder parameters per task. This ensures that task embeddings are accurate representations of tasks. \blue{We use the model of \newcite{CAMTL2021} with latent task embeddings from now on because of its higher performance.}

\subsection{Geometry of Task Embeddings}

Figure \ref{fig:taskviz} displays a 2D projection of the task embeddings  with UMAP \cite{2018arXivUMAP}. 
Some task types, such as sentiment analysis and grammatical properties prediction, form distinct clusters.
Moreover, a PCA\footnote{Unlike UMAP, PCA is a {\it linear} projection of the original space.} projection, which is less readable but provides a more faithful depiction of the global structure, is shown in Appendix \ref{sec:PCA}. This approach allows us to identify linguistic probing tasks (prediction of the number of objects/subjects, prediction of text length, prediction of constituent patterns) as outliers. Since the task embeddings reflect an influence on the conditional adapter, distance from the center can be seen as a way to measure task specificity. Tasks whose embeddings are far from the center need to activate the conditional adapter in a way that is not widely shared and are therefore more specific. 

\begin{table}
\small
\centering
\begin{tabular}{ll}
\toprule
            Task Type &  Position Stability \\
\midrule
              Grammar &     $62.0\pm3.9$ \\
        Acceptability &    $ 57.1\pm0.0$\\
              Emotion &     $47.6\pm2.2$ \\
            Discourse &     $45.7\pm0.0 $\\
                  NLI &     $37.5\pm1.0$ \\
                Other &     $34.8\pm0.7 $\\
 Paraphrase detection &     $31.5\pm13.1$ \\
            Facticity &     $30.0\pm4.7$\\
            \midrule
Random embedding & $1.0\pm0.5$
            
            \\
\bottomrule
\end{tabular}
\caption{Task embeddings position stability within a training run according to task type. As a reference, we provide the expected stability that would be obtained for randomly sampled task embedding positions.
\label{tab:selfstability}
}
\end{table}

\subsection{Stability Analysis}

The appeal of task embeddings relies on the hypothesis that they form similar structures across runs and that each task has a position that does not depend excessively on randomness. In this section, we address these concerns.

%\subsubsection{Sensitivity to Randomness Effects}
\subsubsection{Stability within a Run}
We investigate the sensitivity of task embeddings to initialization and to data sampling order by running the multitask training while assigning $3$ embeddings with different initializations ($ z_{i,1} , z_{i, 2}, z_{i,3}$) to each task instead of $1$. During training, one of the three embeddings is randomly selected for each task training step.

Figure \ref{fig:graph} in Appendix \ref{sec:stabilityviz} displays the task embedding space in this setting. Some task embeddings converge to nearly identical positions (\textit{trec, rotten tomatoes, sst2, mnli}), while the embeddings of other tasks (\textit{boolq, mrpc, answer\_selection\_experiments}) occupy a wider portion of the embedding space. For each task, we compute the rate at which the 10 nearest neighbors \footnote{According to cosine similarity.} of an embedding $z_{i,k}$ contain an embedding of the same task with a different initialization, $z_{i,k'}, k'\neq k$.

The stability rates are reported in Table \ref{tab:selfstability}. The standard deviations (computed across runs) show that sensitivity to random seeds is inherent to the task groups. Some tasks occupy specific regions  in the latent space, while other tasks can lie on multiple positions in a manifold. However, the variability is far from that of random positions.

\begin{table}
\small
\center
\begin{tabular}{ll}

\toprule
            Task Type &    Neighborhood Stability \\
\midrule
              Emotion &$26.3\pm11.2 $\\
              Grammar &$20.2\pm10.4 $\\
        Acceptability &$ 19.4\pm9.1 $\\
 Paraphrase detection &$14.3\pm10.4 $\\
                  NLI &$ 14.1\pm9.5 $\\
            Facticity &$ 13.1\pm8.5 $\\
            Discourse &$ 11.6\pm7.5 $\\
                Other &$ 10.2\pm8.2 $\\
\bottomrule

\end{tabular}
\caption{Task embedding neighborhood stability according to task type.\label{tab:neigh}}
\vspace{-4mm}
\end{table}

\subsubsection{Stability of Task Neighborhood}

\begin{table*}
\begin{small}
\centering
\begin{tabular}{lllllll}
\toprule
                        Features & Domain-Cluster &      Num-Rows & Num-Text-Fields &     Task-Type &   Text-Length &           AVG \\
\midrule
                 n.a. (Majority Class) &           27.8 & \textbf{62.4} &            63.4 &          20.8 &          24.8 &          39.8 \\
Fisher Embedding &           37.7 &          61.3 &            62.2 &          35.7 &          25.6 &          44.5 \\
                 Latent Features &           41.8 &          51.5 &            62.2 &          45.7 &          32.8 &          46.8 \\
  Latent Features$\oplus$TextEmb &           71.5 &          60.4 &            76.3 & \textbf{60.4} &          53.6 &          64.4 \\
                  TextEmb &  \textbf{78.3} &          59.3 &   \textbf{82.3} &          50.6 & \textbf{60.3} & \textbf{66.2} \\
\bottomrule
\end{tabular}

\end{small}
\caption{Accuracy of task aspect classification from task embeddings. $\oplus$ denotes concatenation.}\label{tab:tasktoattribute} 
\end{table*}

\iffalse

\begin{table*}
\begin{small}
\begin{tabular}{lrrrrrr}
\toprule
               Model &    All &  Domain-Cluster &  Num-Examples &  Num-Text-Fields &  Task-Type &  Text-Length \\
\midrule
      Average Task Embedding &  1.000 &             - &       - &              - &        - &          - \\
 KNN Regression &  0.955 &           0.986 &     1.056 &            1.000 &      0.953 &        1.021 \\
               Ridge Regression &  0.907 &           0.956 &     0.997 &            1.005 &      0.918 &        0.988 \\
\bottomrule
\end{tabular}
\end{small}
\caption{Mean squared error (MSE) of embedding regression from aspects. To normalize the reported MSE values, we divide them by the MSE of average task embedding prediction.
\label{tab:offline}}
\end{table*}

\fi

We study the neighborhood of each task embedding. Following \newcite{antoniak-mimno-2018-evaluating}, we define the stability rate for a task embedding as the average overlap rate (according to the Jaccard metric) of the  neighborhoods.

Given two spaces A and B from different runs and a task $\T_i$ , we define the neighborhood of $\T_i$ in A as the top 10 closest other tasks according to cosine similarity. We also compute the neighborhood of $\T_i$ in B. 
We report the results according to task type in Table \ref{tab:neigh}. The results show that the global structure of the space can change and that task type influences the neighborhood stability.

\textbf{RQ1} can be answered with a distinction on the task type. The position of a task embedding within a run is relatively robust to randomness. Across runs, the organization of the task embedding space may vary. In both cases, lower-level tasks, such as grammar, acceptability, and emotion tasks, exhibit the most consistent structure.

\subsection{Probing Task Embeddings for Task Aspects}
\label{sec:probingforattributes}

 \begin{table*}
\centering
\begin{small}

\begin{tabular}{lllllllll}
\toprule
                                           &           CoLa &           SST2 &           MRPC &            QQP &           MNLI &           QNLI &            RTE &            AVG \\
\midrule
                   Single-Task Full-Fine-Tuning (Supervised)    &           79.2 &           93.1 &           75.5 &           84.7 &           80.9 &           88.9 &           47.3 &                 78.5 \\
                   \midrule
                Same Task-Type Full Fine-Tuning &           73.5 &  \textbf{93.6} &           68.8 &           55.3 &  \textbf{72.7} &           51.5 &  \textbf{70.2} &           69.4 \\
             \midrule
                 Same Task-Type Task Embeddings &  \textbf{76.7} &           91.4 &           67.6 &           57.0 &           67.0 &           53.8 &           64.0 &           68.2 \\
                  Offline Task Embedding Ridge  Regression &           76.2 &           92.0 &           67.6 &           61.6 &           71.7 &           53.8 &           68.6 &           70.2 \\
                  
           Features-Aware Task Embeddings - TextEmb  \cite{vu-etal-2020-exploring}&           75.4 &  90.2 &  70.2 &  53.9 &           66.5 &  57.3 &          65.8 & 	68.1 \\

           Features-Aware Task Embeddings - Aspects (ours) &           75.4 &  90.0 &  \textbf{70.4} &  \textbf{71.1} &           66.2 &  \textbf{56.2} &           63.7 & \textbf{70.4} \\

\bottomrule
\end{tabular}

\end{small}
\caption{Zero-Shot (ZS) accuracy on GLUE tasks after training on MetaEval while excluding GLUE tasks (ME$\setminus$G). As a reference, we also provide results with supervision on each evaluated task with the setup from section \ref{sec:setup}. The Same-Task-Type is the baseline, where for each task, RoBERTa is fine-tuned on (ME$\setminus$G) same-type tasks while sharing label weights.
The next methods use task embedding prediction via either offline or online regression, as described in section \ref{sec:tep}. \label{tab:glueinference}}
\end{table*}

We now use the task embeddings to investigate which task aspects influence the NLP models. Prior work developed a probing methodology to interpret the content of \textit{text} embeddings. \newcite{conneau-etal-2018-cram} selected an array of text aspects to see if they were contained in the text embedding. These aspects include text length, word content, the number of subjects and objects, the tense, natural word order, and syntactic properties.

To derive analogous \textit{task} aspects $\Lambda_i$, we model a task as a collection of text examples with labels. We propose as aspects the number of text examples, the number of text fields per example, and the type of task. We also include basic properties derived from the text of the examples, namely, the median text length and the domain. 

\paragraph{Num-Examples} represents the number of training examples for a task. We discretize this value into 4 quartiles\footnote{We experimented with finer quantizations, but they led to excessive sparsity.} computed across all tasks.
\paragraph{Num-Text-Fields} is equal to 2 in sentence-pair classification tasks (e.g., NLI or paraphrase detection) and equal to 1 in single-sentence classification tasks (e.g., standard sentiment analysis).
\paragraph{Domain-Cluster} is a representation of the domain of the input text of a task. Following \cite{sia-etal-2020-tired}, we represent the text of each task by the average spherical embedding \cite{meng2019spherical}.
The domain of each task is represented by the average of the text embeddings of its examples. We then perform clustering across all task domains  to reduce the dimensionality of the domain representation. We use Gaussian mixture model soft clustering and represent the domain by 8 cluster activations.\footnote{The number of clusters was selected with the elbow method.}

\paragraph{Text-Length} represents the length of the input examples (and the sum of input lengths when there are two inputs). We discretize this value into 4 quartiles computed across all tasks.
\paragraph{Task-Type} is the type of task, selected from $\{$ $\textsc{Acceptability}$,
$\textsc{Discourse}$,
$\textsc{Emotion}$,
$\textsc{Facticity}$,
$\textsc{Grammar}$,
$\textsc{NLI}$,
$\textsc{Paraphrase detection}$,
$ \textsc{Other}$ $\}$. 

Note that the above aspects do not rely on annotated data (only on the input text, sizes, and task type). We use a logistic regression classifier with Scikit-Learn \cite{scikit-learn} default parameters\footnote{Release 0.24.1; deviation from the default parameters did not lead to a significant improvement. We also experimented with gradient boosting trees and KNN classifiers with no improvement.} to learn to predict the aspects from task embeddings. Table \ref{tab:tasktoattribute} displays the classification accuracy for each aspect obtained by performing cross-validation with a leave-one-out split.

The number of training examples is limited to the number of tasks, which prevents high accuracy. However, our results address \textbf{RQ2} by showing that a simple linear probe can still capture the domain, the task type, and the length of the input.
Fisher Embeddings perform poorly, but \newcite{vu-etal-2020-exploring} explain that, unlike other methods, the Fisher embeddings do not lie in a comparable space. TextEmb performs surprisingly well in these probing tasks, however, it does not fully capture the task type, since concatenation with the latent embedding improves the classification of this aspect. This can explain the relatively low performance of TextEmb in table \ref{tab:metaevalacc}.
The task embeddings do not seem to accurately capture the difference between single sentence or sentence pair tasks, except for TextEmb, which is sensitive to separator tokens.
\subsection{
Task Embedding Regression \label{sec:tep}}

We now address the prediction of task embeddings from the previously defined aspects. 
We use task embeddings $z_i$ trained with the MetaEval multitask setup and then train a regression model  to predict the task embeddings from the task aspects $ \Lambda_{\T_i}$ or TextEmb.
\vspace{-0.5mm}
\begin{equation}
    \hat{z}_i=\text{Regression}([a, a\in \Lambda_{\T_i}])
\end{equation}
%We propose two evaluations, intrinsic and extrinsic. In our first evaluation, we directly measure the regression error, which allows us to measure how well trained task embeddings can be recovered on the basis of aspects alone. Table \ref{tab:offline} shows the error with two regression models. The  ridge regression model outperforms neighborhood-based regression (KNN), which shows that relevant aspects can be abstracted from the embeddings even on $\approx 100$ examples.

To evaluate task embedding regression, we exclude GLUE tasks from MetaEval during the multitask conditional adapter training. We now share the label names across tasks during the multitask training to enable zero-shot inference. Then, we estimate task embeddings for the GLUE classification tasks from the aspects via logistic regression. We propose two different techniques for task embedding regression:
\paragraph{Offline Task Embedding Regression}

We first perform multitask training, then train a regression model to estimate task embeddings from a set of aspects.
One advantage of this technique is that it allows the use of any aspect after  multitask training. However, the model has to learn this relationship from only 100 examples since an example is a task.

\paragraph{Features-Aware Task Embeddings}
We propose another variation, where we perform multitask training and the regression of embeddings jointly.
Instead of using only a latent task embedding $z_i$ for each task $\T_i$, we add it to a projection of the input features $\phi_i$, which can be either a concatenation of all aspects\footnote{Using one-hot representations.}, or TextEmb.  The task embedding modulating the adapters is then $z_{i} + W_{\phi}\phi_i$. An unseen task $\T_i$ can be represented by  the projection from aspect embeddings augmented with the average latent task embedding. 

As another baseline, we propose the \textbf{Same-Task-Type Full Fine-Tuning} of a RoBERTa model. For each GLUE task, we fine-tune the model on all MetaEval tasks of the same task type \cite{mou-etal-2016-transferable} while excluding GLUE tasks. For instance, to derive predictions on RTE, we fine-tune a RoBERTa model on all NLI tasks of MetaEval that are not in GLUE while sharing the labels.
We also report the results of supervised RoBERTa models trained on each GLUE task with the hyperparameters described in section \ref{sec:setup}.

Table \ref{tab:glueinference} reports the GLUE accuracy under both settings. Task embedding regression improves the average accuracy compared to that of the Same-Task-Type RoBERTa baseline. Learning aspect embeddings during multitask training leads to an improved average result, but most of the gain over the baseline is achieved via offline regression.
\blue{Using TextEmb as features performs worse than using latent embeddings which indicates that TextEmb does not capture enough important information about the task type\footnote{Fisher task embeddings led to lower accuracies.}.}
Finally, averaging the task embeddings of the same-type tasks leads to the worst results, which confirms the need to combine multiple aspects of a task for task embedding prediction. These findings address \textbf{RQ3} and establish task embeddings as a viable gateway for zero-shot transfer of NLP tasks based on task attributes.

\section{Conclusion}
We proposed a framework for the analysis and prediction of task embeddings in NLP. We showed that the task embedding space exhibits a consistent structure but that there are individual variations according to task type. Furthermore, we have demonstrated that task embeddings can be predicted based on task aspects. Since the task embedding leads
to a model,  model manipulation can be performed according to desirable aspects for zero-shot prediction. Future work can consider new task aspects for model manipulation, such as undesirable associations language.

\section{Acknowledgements}
This work is part of the CALCULUS project, which is
funded by the ERC Advanced Grant H2020-ERC-2017.
ADG 788506\footnote{\url{https://calculus-project.eu/}}.

\nocite{HendrycksG16}

\bibliography{main, references}
\bibliographystyle{lrec2022-bib}

\clearpage
\onecolumn
\appendix

\section{Fisher Information Task Embedding\label{sec:fisher}}

The average of text embeddings for all samples of  a task can be used as a task embedding, but they ignore the labels entirely. \newcite{achille2019task2vec} propose a task embedding based on the influence of a task training objective on network weights. To do so, they use an empirical Fisher information estimate of a fine-tuned network as a task embedding. Fisher information captures the influence of model parameters or activations on the loss function. For BERT-based models, \newcite{vu-etal-2020-exploring} suggest the use of $h_{\text{[CLS]}}$ token activation and to only consider the diagonal information of the Fisher matrix, which is the expected variance of the gradients of the log-likelihood with respect to activations. Activation dimensions that are important for a task will have a high fisher information. Since similar tasks should use similar features, Fisher information of the activations can capture useful task representations. As suggested by \newcite{vu-etal-2020-exploring}, we perform a fine-tuning for each task before task embedding computation with the setting of section \ref{sec:targettask}.  We then compute the empirical Fisher information embeddings for a task $i$ as follows: 

\begin{equation}
    F_{\theta}(\T_i) = \frac{1}{n} \sum_{k=1}^{n}
    (\nabla_{\theta} \log P_{\theta} (y_{k},  x_{k}) )^{2}
    \end{equation}
Where $n$ is the number of training samples.
$\theta$ can be the full network or any activation but here we use the $h_{\text{[CLS]}}$ activation, which achieved the best results in our section \ref{sec:targettask} experiment.

\section{List of Tasks}
\label{sec:tasklist}
\begin{footnotesize}
\begin{longtable}{lll}
\toprule
                                    Dataset &                                             Labels &   Splits Sizes \\
\midrule
\endhead
\midrule
\multicolumn{3}{r}{{Continued on next page}} \\
\midrule
\endfoot

\bottomrule
\endlastfoot
                        health\_fact/default &                   [false, mixture, true, unproven] &      10k/1k/1k \\
                         ethics/commonsense &                         [acceptable, unacceptable] &      14k/4k/4k \\
                          ethics/deontology &                         [acceptable, unacceptable] &      18k/4k/4k \\
                             ethics/justice &                         [acceptable, unacceptable] &      22k/3k/2k \\
                      ethics/utilitarianism &                         [acceptable, unacceptable] &      14k/5k/4k \\
                              ethics/virtue &                         [acceptable, unacceptable] &      28k/5k/5k \\
                        discovery/discovery &  [[no-conn], absolutely,, accordingly, actually... &     2M/87k/87k \\
                               ethos/binary &                      [no\_hate\_speech, hate\_speech] &            998 \\
                            emotion/default &        [sadness, joy, love, anger, fear, surprise] &      16k/2k/2k \\
                      hate\_speech18/default &                 [noHate, hate, idk/skip, relation] &            11k \\
                    pragmeval/verifiability &     [experiential, unverifiable, non-experiential] &      6k/2k/634 \\
                  pragmeval/emobank-arousal &                                        [low, high] &     5k/684/683 \\
                      pragmeval/switchboard &  [Response Acknowledgement, Uninterpretable, Or... &     19k/2k/649 \\
         pragmeval/persuasiveness-eloquence &                                        [low, high] &      725/91/90 \\
                             pragmeval/mrda &  [Declarative-Question, Statement, Reject, Or-C... &      14k/6k/2k \\
                              pragmeval/gum &  [preparation, evaluation, circumstance, soluti... &     2k/259/248 \\
                         pragmeval/emergent &                          [observing, for, against] &     2k/259/259 \\
         pragmeval/persuasiveness-relevance &                                        [low, high] &      725/91/90 \\
       pragmeval/persuasiveness-specificity &                                        [low, high] &      504/62/62 \\
          pragmeval/persuasiveness-strength &                                        [low, high] &      371/46/46 \\
                pragmeval/emobank-dominance &                                        [low, high] &     6k/798/798 \\
              pragmeval/squinky-implicature &                                        [low, high] &     4k/465/465 \\
                          pragmeval/sarcasm &                                    [notsarc, sarc] &     4k/469/469 \\
                pragmeval/squinky-formality &                                        [low, high] &     4k/453/452 \\
                             pragmeval/stac &  [Comment, Contrast, Q\_Elab, Parallel, Explanat... &      11k/1k/1k \\
                             pragmeval/pdtb &  [Synchrony, Contrast, Asynchronous, Conjunctio... &      13k/1k/1k \\
       pragmeval/persuasiveness-premisetype &  [testimony, warrant, invented\_instance, common... &      566/71/70 \\
          pragmeval/squinky-informativeness &                                        [low, high] &     4k/465/464 \\
         pragmeval/persuasiveness-claimtype &                              [Value, Fact, Policy] &      160/20/19 \\
                  pragmeval/emobank-valence &                                        [low, high] &     5k/644/643 \\
                           hope\_edi/english &        [Hope\_speech, Non\_hope\_speech, not-English] &         23k/3k \\
                            snli/plain\_text &               [entailment, neutral, contradiction] &   550k/10k/10k \\
                         paws/labeled\_final &                                             [0, 1] &      49k/8k/8k \\
                            imdb/plain\_text &                                         [neg, pos] &    50k/25k/25k \\
        crowdflower/sentiment\_nuclear\_power &  [Neutral / author is just sharing information,... &            190 \\
           crowdflower/tweet\_global\_warming &                                          [Yes, No] &             4k \\
              crowdflower/airline-sentiment &                      [neutral, positive, negative] &            15k \\
            crowdflower/corporate-messaging &           [Information, Action, Exclude, Dialogue] &             3k \\
                  crowdflower/economic-news &                                [not sure, yes, no] &             8k \\
       crowdflower/political-media-audience &                           [constituency, national] &             5k \\
           crowdflower/political-media-bias &                                [partisan, neutral] &             5k \\
        crowdflower/political-media-message &  [information, support, policy, constituency, p... &             5k \\
                   crowdflower/text\_emotion &  [sadness, empty, relief, hate, worry, enthusia... &            40k \\
                                emo/emo2019 &                        [others, happy, sad, angry] &         30k/6k \\
                                  glue/cola &                         [unacceptable, acceptable] &       9k/1k/1k \\
                                  glue/sst2 &                               [negative, positive] &     67k/2k/872 \\
                                  glue/mrpc &                       [not\_equivalent, equivalent] &      4k/2k/408 \\
                                   glue/qqp &                         [not\_duplicate, duplicate] &  391k/364k/40k \\
                                  glue/mnli &               [entailment, neutral, contradiction] &   393k/10k/10k \\
                                  glue/qnli &                       [entailment, not\_entailment] &     105k/5k/5k \\
                                   glue/rte &                       [entailment, not\_entailment] &      3k/2k/277 \\
                                  glue/wnli &                       [not\_entailment, entailment] &     635/146/71 \\
                                    glue/ax &               [entailment, neutral, contradiction] &             1k \\
          yelp\_review\_full/yelp\_review\_full &        [1 star, 2 star, 3 stars, 4 stars, 5 stars] &       650k/50k \\
      blimp\_classification/syntax\_semantics &                         [acceptable, unacceptable] &            26k \\
      blimp\_classification/syntax+semantics &                         [acceptable, unacceptable] &             2k \\
            blimp\_classification/morphology &                         [acceptable, unacceptable] &            36k \\
                blimp\_classification/syntax &                         [acceptable, unacceptable] &            52k \\
             blimp\_classification/semantics &                         [acceptable, unacceptable] &            18k \\
                 recast/recast\_kg\_relations &                                 [1, 2, 3, 4, 5, 6] &     22k/2k/761 \\
                         recast/recast\_puns &                           [not-entailed, entailed] &      14k/2k/2k \\
                   recast/recast\_factuality &                           [not-entailed, entailed] &      38k/5k/4k \\
                      recast/recast\_verbnet &                           [not-entailed, entailed] &     1k/160/143 \\
                   recast/recast\_verbcorner &                           [not-entailed, entailed] &   111k/14k/14k \\
                          recast/recast\_ner &                           [not-entailed, entailed] &   124k/38k/36k \\
                    recast/recast\_sentiment &                           [not-entailed, entailed] &     5k/600/600 \\
             recast/recast\_megaveridicality &                           [not-entailed, entailed] &       9k/1k/1k \\
                            ag\_news/default &                [World, Sports, Business, Sci/Tech] &        120k/8k \\
                           super\_glue/boolq &                                      [False, True] &       9k/3k/3k \\
                              super\_glue/cb &               [entailment, contradiction, neutral] &     250/250/56 \\
                             super\_glue/wic &                                      [False, True] &      5k/1k/638 \\
                             super\_glue/axb &                       [entailment, not\_entailment] &             1k \\
                             super\_glue/axg &                       [entailment, not\_entailment] &            356 \\
 ade\_corpus\_v2/Ade\_corpus\_v2\_classification &                             [Not-Related, Related] &            24k \\
                            tweeteval/emoji &  [\_red\_heart\_, \_smiling\_face\_with\_hearteyes\_, \_... &     50k/45k/5k \\
                             tweeteval/hate &                                   [not-hate, hate] &       9k/3k/1k \\
                            tweeteval/irony &                                 [non\_irony, irony] &     3k/955/784 \\
                        tweeteval/offensive &                         [not-offensive, offensive] &     12k/1k/860 \\
                        tweeteval/sentiment &                      [negative, neutral, positive] &     46k/12k/2k \\
                           tweeteval/stance &                      [negative, neutral, positive] &      3k/1k/294 \\
                               trec/default &  [manner, cremat, animal, exp, ind, gr, title, ... &         5k/500 \\
                   yelp\_polarity/plain\_text &                                             [1, 2] &       560k/38k \\
                    rotten\_tomatoes/default &                                         [neg, pos] &       9k/1k/1k \\
                            anli/plain\_text &               [entailment, neutral, contradiction] &   100k/45k/17k \\
                               liar/default &  [false, half-true, mostly-true, true, barely-t... &      10k/1k/1k \\
              linguisticprobing/subj\_number &                                          [NN, NNS] &      82k/8k/8k \\
               linguisticprobing/obj\_number &                                          [NN, NNS] &      80k/8k/8k \\
             linguisticprobing/past\_present &                                       [PAST, PRES] &      86k/9k/9k \\
          linguisticprobing/sentence\_length &                                 [0, 1, 2, 3, 4, 5] &      87k/9k/9k \\
         linguisticprobing/top\_constituents &  [ADVP\_NP\_VP\_., CC\_ADVP\_NP\_VP\_., CC\_NP\_VP\_., IN... &      70k/7k/7k \\
               linguisticprobing/tree\_depth &  [depth\_5, depth\_6, depth\_7, depth\_8, depth\_9, ... &      85k/9k/9k \\
   linguisticprobing/coordination\_inversion &                                             [I, O] &   100k/10k/10k \\
              linguisticprobing/odd\_man\_out &                                             [C, O] &      83k/8k/8k \\
             linguisticprobing/bigram\_shift &                                             [I, O] &   100k/10k/10k \\
             snips\_built\_in\_intents/default &  [ComparePlaces, RequestRide, GetWeather, Searc... &            328 \\
            amazon\_polarity/amazon\_polarity &                               [negative, positive] &        4M/400k \\
                        winograd\_wsc/wsc285 &                                             [0, 1] &            285 \\
                        winograd\_wsc/wsc273 &                                             [0, 1] &            273 \\
                              hover/default &                         [NOT\_SUPPORTED, SUPPORTED] &      18k/4k/4k \\
                      dbpedia\_14/dbpedia\_14 &  [Company, EducationalInstitution, Artist, Athl... &       560k/70k \\
                    onestop\_english/default &                                    [ele, int, adv] &            567 \\
                   movie\_rationales/default &                                         [NEG, POS] &     2k/200/199 \\
                            hans/plain\_text &                       [entailment, non-entailment] &        30k/30k \\
               sem\_eval\_2014\_task\_1/default &               [NEUTRAL, ENTAILMENT, CONTRADICTION] &      5k/4k/500 \\
                    eraser\_multi\_rc/default &                                      [False, True] &      24k/5k/3k \\
         selqa/answer\_selection\_experiments &                                             [0, 1] &     66k/19k/9k \\
                         scitail/tsv\_format &               [entailment, neutral, contradiction] &      23k/2k/1k \\
\end{longtable}
\label{taskslist}
\end{footnotesize}
\clearpage
\section{Task Embedding Stability\label{sec:stabilityviz}}
\begin{figure*}[h!]

  \centering
  \vspace{-0.1cm}
\includegraphics[trim={0cm 0 0cm 0},clip,
width =1.0\textwidth,]{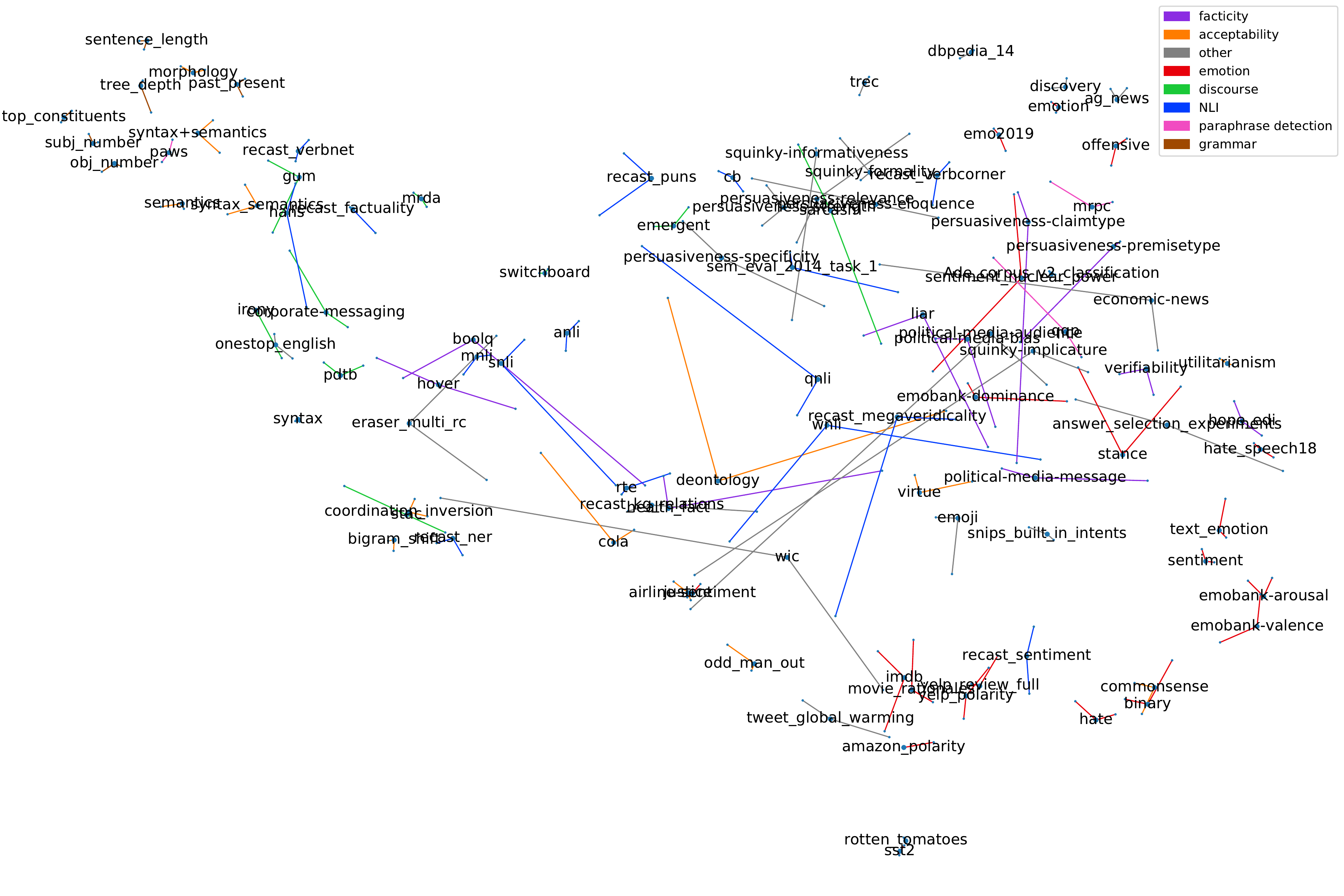}
\begin{normalsize}
  \caption{UMAP Visualization of task embeddings when each task is attributed 3 task embeddings. For each task, we position the task name at the centroid of the three embeddings and represent edges between the centroid and the two other embeddings.}
\label{fig:graph} 
\end{normalsize}

\end{figure*}
\clearpage
\section{PCA Visualization of Task Embeddings\label{sec:PCA}}
\begin{figure*}[h!]

  \centering
  \vspace{-0.1cm}
\includegraphics[trim={3cm 2.5cm 3cm 3cm},clip,
width =1.0\textwidth,]{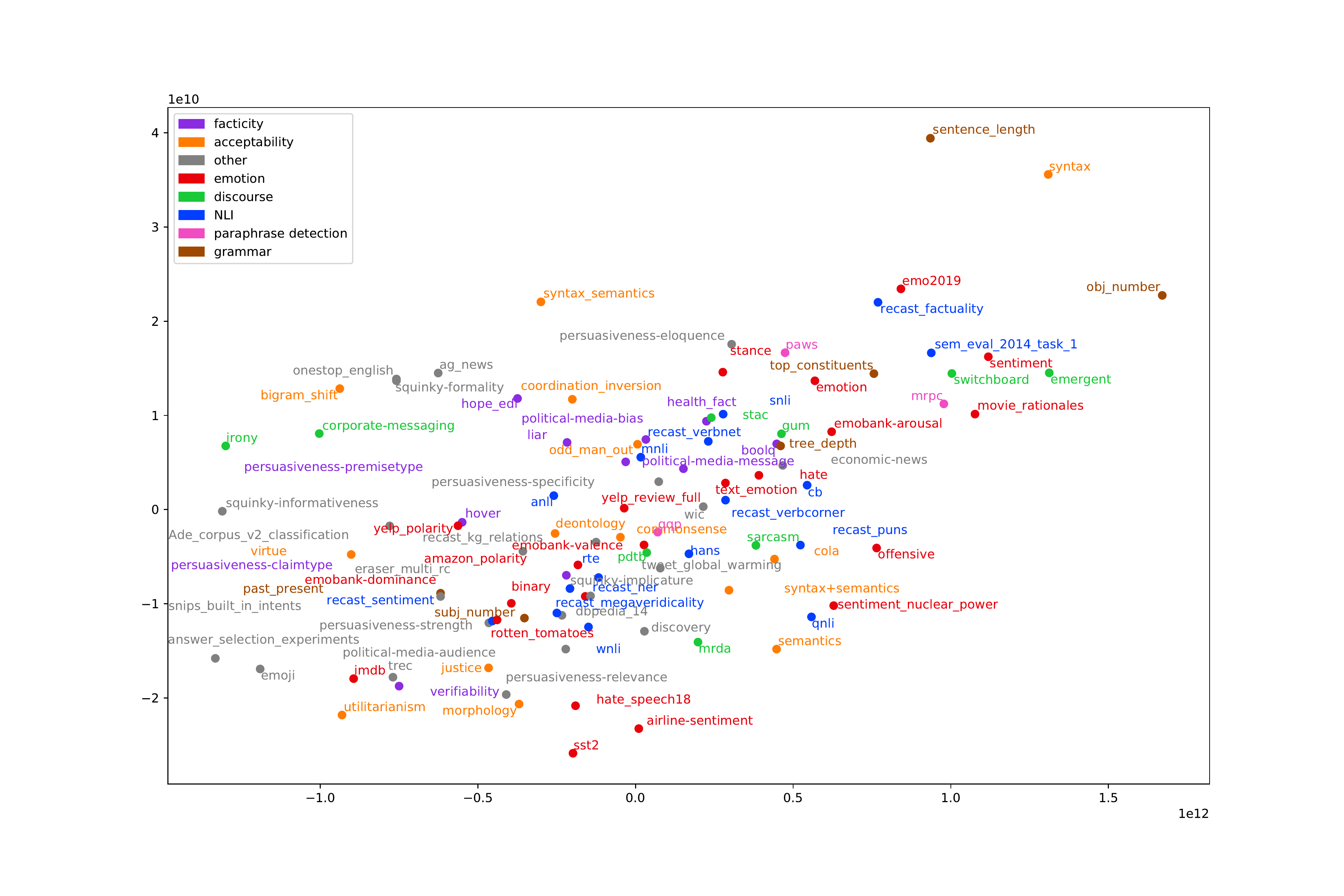}
\begin{normalsize}
  \caption{PCA Visualization of task embeddings.}
\label{fig:graphpca} 
\end{normalsize}

\end{figure*}

\end{document}